%% file: FedHLCR.tex
\documentclass{article}

\usepackage{PRIMEarxiv}

\usepackage[utf8]{inputenc} 
\usepackage[T1]{fontenc}    
\usepackage{hyperref}       
\usepackage{url}            
\usepackage{booktabs}       
\usepackage{amsfonts}       
\usepackage{nicefrac}       
\usepackage{microtype}      
\usepackage{lipsum}
\usepackage{fancyhdr}       
\usepackage{graphicx}       
\usepackage{authblk}
\graphicspath{{media/}}     

\usepackage{amsmath}
\usepackage{graphicx}
\usepackage{algorithm}
\usepackage[italicComments=false,commentColor=black]{algpseudocodex}
\usepackage{comment}
\usepackage{caption}
\usepackage{ntheorem}
\usepackage{enumitem}

\title{Federated Latent Class Regression for \\ Hierarchical Data}

\author[1]{Bin Yang}
\author[2]{Thomas Carette}
\author[1]{Masanobu Jimbo}
\author[1]{Shinya Maruyama}
\affil[1]{Sony Group Corporation, Tokyo, Japan}
\affil[2]{Sony Europe B.V., Brussels, Belgium}

\begin{document}
\maketitle

\begin{abstract}
\input abstract.tex
\end{abstract}

\section{Introduction}\label{sec.Introduction}
\input introduction.tex

\section{Related Works}\label{sec.related_work}
\input related_work.tex

\input algorithm.tex

\section{Experiments}\label{sec.experiment}
\input experiment.tex

\section{Conclusion}
\input conclusion.tex

\section*{Broader Impact}
\input broader_impact.tex

\bibliographystyle{unsrt}  
\bibliography{reference}

\input appendix.tex

\end{document}

%% file: abstract.tex
Federated Learning (FL) allows a number of agents to participate in training a global machine learning model without disclosing locally stored data.
Compared to traditional distributed learning, the heterogeneity (non-IID) of the agents slows down the convergence in FL.
Furthermore, many datasets, being too noisy or too small, are easily overfitted by complex models, such as deep neural networks.
Here, we consider the problem of using FL regression on noisy, hierarchical and tabular datasets in which user distributions are significantly different.
Inspired by Latent Class Regression (LCR), we propose a novel probabilistic model, Hierarchical Latent Class Regression (\textsc{HLCR}), and its extension to Federated Learning, \textsc{FedHLCR}.
\textsc{FedHLCR} consists of a mixture of linear regression models, allowing better accuracy than simple linear regression, while at the same time maintaining its analytical properties and avoiding overfitting. 
Our inference algorithm, being derived from Bayesian theory, provides strong convergence guarantees and good robustness to overfitting.
Experimental results show that \textsc{FedHLCR} offers fast convergence even in non-IID datasets.

%% file: introduction.tex
Machine learning technology has been introduced into a variety of services and has become a cornerstone of industrial society.
Machine learning models require a large amount of data in most cases, and sometimes these data contain private information.
Federated Learning (FL) is a technique to lessen the risk of privacy leakage while preserving the benefits of distributed machine learning systems~\cite{Konecny16, McMahan2017, Bonawitz19},
in which agents, such as smartphones and automobiles, cooperatively train a machine learning model by communicating with a central server without disclosing private information.

While FL has received huge attention in recent years and has already been deployed in several services~\cite{hard2018federated,wwdc2019,leroy2019federated}, there are still many technical challenges~\cite{mcmahan2021advances}. One of the biggest challenges is data heterogeneity among agents.
The datasets of FL are constituted of the local data of a set of agents, and their distributions are typically different, i.e. non-\emph{Identically and Independently Distributed} (non-IID). 
The Stochastic Gradient Descent (SGD) method~\cite{Robbins51, Kiefer52} has been central in scaling learning algorithms to large models, such as Deep Neural Networks (DNN).
Although it is well known that FedAve, the most basic federated SGD algorithm, converges under certain conditions, it is still difficult to guarantee accuracy and convergence in general~\cite{zhao2018federated,li2020convergence}.
Various approaches, including some clustering techniques~\cite{sattler2020clustered,ghosh2020efficient}, have already been proposed to tackle this problem.
However, those methods do not fit the case where each agent has data which is heterogeneously distributed. 
To explain this aspect, let us introduce a specific use case.
A smartphone system decides whether to connect to a specific Wi-Fi access point within the signal range by predicting the connection quality.
Even for the same smartphone, different Wi-Fi access points result in different connection quality due to differences in the properties of access points.
In other words, the connection quality depends on not only the devices but also the access points.
In the above example, each agent (smartphone) contains data in a set of entities (Wi-Fi access points) and each entity contains a set of connection records.
We call this kind of structure a \emph{hierarchical data structure}.
A hierarchical data structure is encountered in many contexts, such as patient-hospital data in medical sciences and customer-product data in recommendation systems.
Developing a single model to perform predictions on a non-IID hierarchical dataset is complex.
Especially in the case of neural networks, for which the generalization properties are not deeply understood, we have a risk of overfitting issues~\cite{Neyshabur17, Nagarajan19}.
Furthermore, it can lead to prohibitive computational costs on edge devices, as well as slow convergence, leading in turn to even higher costs in communication and edge computing.

To handle non-IID hierarchical data efficiently, we built a pure Bayesian inference model called Hierarchical Latent Class Regression (HLCR) and FedHLCR, which is an extension of HLCR to FL, by combining linear regressors in a hierarchical mixture manner inspired by Latent Dirichlet Allocation (LDA), which works well for hierarchical data. 
In addition, we propose an optimization algorithm applying Collapsed Gibbs Sampling (CGS) and guarantee significant acceleration in the convergence of HLCR. 
The key point is to cluster data in each agent based on the entity and to train a simple model per cluster without disclosing any private information.
In this paper, we only focus on the mixture of linear models since it effectively avoids the overfitting problem.

{\bf Our contributions:}
To the best of our knowledge, this is the first research on the regression problem for hierarchical data in a Federated Learning setting.
\begin{itemize}
  \item We establish a purely probabilistic mixture model, called Hierarchical Latent Class Regression (HLCR), which is mixed by {\sl linear regression model}s. The hierarchical structure allows HLCR to handle hierarchical data very well.
\item We propose an efficient Collapsed Gibbs Sampling algorithm for inferring an HLCR model with fast convergence.
\item We extend the Collapsed Gibbs Sampling algorithm of HLCR to the Federated Learning (FedHLCR), preventing sensitive information from each agent from being disclosed.
\end{itemize}


%% file: related_work.tex
{\bf Federated Learning:} FL~\cite{Konecny16,McMahan2017} is a powerful method to train a machine learning model in the distributed setting; however, it has a variety of technical challenges, and non-IID data distribution among agents is a central problem.
Many researchers have studied the performance of FedAve and similar FL algorithms, and extensions of SGD, on non-IID settings~\cite{Lietal20,zhao2018federated,li2020convergence,karimireddy2020scaffold,ahmed2019,wu2021node}.
\cite{zhao2018federated} shows that the accuracy and convergence speed of FedAve are reduced on non-IID data compared with IID data, and \cite{Lietal20} proposes a method, called FedProx, which adds a proximal term to the objective function to prevent each local model from overfitting data in each agent.
Several studies suggest to use multiple models to address the non-IID problem~\cite{smith2017MOCHA,sattler2020clustered,zantedeschi2020fully,fallar2020,jiang2019improving,mansour2020three,ghosh2020efficient}, and some researchers adopt multitask learning~\cite{smith2017MOCHA,sattler2020clustered,zantedeschi2020fully} or meta learning~\cite{fallar2020,jiang2019improving} to deal with multiple targets.
Another approach applies the latent class problem to FL for clustering agents~\cite{ghosh2020efficient} and it assumes that each agent has data in one entity, but our method is more general in that each agent is considered to have data in multiple entities. 

{\bf Mixture Models:} Probabilistic Mixture Models have been studied for more than 100 years \cite{newcomb1886generalized,10.2307/90707}. The classical Mixture Model is the Gaussian Mixture Model (GMM) \cite{bishop2006pattern}, which can be inferred with an expectation–maximization (EM) algorithm and has been widely used in clustering. Our proposal in this paper is inspired by two classes of Mixture Models. The first is Latent Class Regression (LCR) \cite{wedel1994review,Grun2008}, or Mixture of Regression.
The second class is topic models which generally handle document-word data,
including Latent Dirichlet Allocation (LDA) \cite{blei2003latent}
and its variants \cite{wallach2009rethinking,blei2006correlated,teh2006hierarchical}.

%% file: algorithm.tex
\section{Mixture Models}\label{sec.mixture_model}
In statistics, a mixture model is a probabilistic model composed of $K$ simple models, which are also called clusters. Each of these models has the same probability distribution but with different parameters. The probability density function of the $k$-th model is denoted as $p(\cdot | \boldsymbol{\varphi}_k)$, where $\boldsymbol{\varphi}_k$ is the parameter of the $k$-th model. Hence, the density of the mixture model can be denoted as $\Sigma_{k=1}^K \theta_k p(\cdot | \boldsymbol{\varphi}_k)$, where the mixture weight $\boldsymbol{\theta} = (\theta_1, \theta_2, \cdots, \theta_K)$ is a $K$-simplex; i.e., $\theta_k \geq 0$, $\Sigma_{k=1}^K \theta_k = 1$. In a Bayesian setting, the mixture weights and parameters are regarded as unknown random variables and will be inferred from observations. Each observation generated from mixture model $\Sigma_{k=1}^K \theta_k p(\cdot | \varphi_k)$ can be equivalently generated by the following two steps: 1) a latent cluster label $z_i$ is sampled from categorical distribution with parameter $\boldsymbol \theta = (\theta_1, \theta_2, \dots, \theta_K)$; then 2) an observation is sampled from the corresponding model $p(\cdot | \varphi_{z_i})$.

{\bf Notation:} In this paper, we use lowercase letters for scalars, bold lowercase letters for vectors, and bold uppercase letters for matrices. The set $\{1,2,\cdots,K\}$ is denoted as $[K]$. We summarize all notations used in the paper in Table \ref{tb:notations}.

\begin{table}[htb]
  \begin{tabular}{rl|rl} \hline
    Notation & Description & Notation & Description \\ \hline
    $F \geq 1$ & integer, the dimension of features & $K \geq 1$ & integer, the number of clusters \\
    $\alpha, \beta, \delta, \sigma, \eta$ & positive scalar, prior of distribution & $\boldsymbol{\theta}$, $\boldsymbol{\psi}$, $\boldsymbol{\theta}_i$ & $K$-simplex, categorical distribution \\
    $\mathbf{w}_k$ & $F$-dimensional coefficient vector & $\boldsymbol{\varphi}_k$ & simplex, categorical distribution \\
    $N, N_i, N_{ij}$ & the number of data & $N^{(k)}, N_i^{(k)}$ & the number of data in cluster $k$ \\
    $x_{ij}$ & one observed record & $z_i, z_{ij}$ & cluster label ($z_* \in [K]$) \\
    $\mathbf{z}$ & cluster labels of all data & $\mathbf{z}_{\setminus ij}$ & cluster labels of data except $z_{ij}$ \\
    $\mathbf{x}_i$, $\mathbf{x}_{ijn}$ & feature (column) vector of a record & $y_i$, $y_{ijn}$ & target value of a record \\
    $\mathbf{X}$ & feature matrix of all records & $\mathbf{y}$ & target vector of all records \\
    $\mathbf{X}_{ij}$ & feature matrix of records $\mathbf{x}_{ij*}$; & $\mathbf{y}_{ij}$ & target vector of records $y_{ij*}$; \\
     & i.e., $(\mathbf{x}_{ij1}, \mathbf{x}_{ij2}, \cdots, \mathbf{x}_{ijN_{ij}})^\top$ & & i.e., $(y_{ij1}, y_{ij2}, \cdots, y_{ijN_{ij}})^\top$ \\
    $\mathbf{X}_{\setminus ij}$ & feature matrix of records except $\mathbf{X}_{ij}$ & $\mathbf{y}_{\setminus ij}$ & target vector of records except $\mathbf{y}_{ij}$ \\
    $\mathbf{X}^{(k)}$ & feature matrix of records in cluster $k$ & $\mathbf{y}^{(k)}$ & target vector of records in cluster $k$ \\
    $\mathbf{A}_n, \mathbf{D}^{(k)}$ & $(F \times F)$-matrix & $\mathbf{b}_n, \mathbf{c}^{(k)}$ & $F$-vector \\
    $\mathbf{D}$ & $(\mathbf{D}^{(1)}, \mathbf{D}^{(2)}, \cdots, \mathbf{D}^{(K)})$ & $\mathbf{c}$ & $(\mathbf{c}^{(1)}, \mathbf{c}^{(2)}, \cdots, \mathbf{c}^{(K)})$ \\ \hline
  \end{tabular}
  \caption{Notations}
  \label{tb:notations}
\end{table}


\begin{figure}[H]
    \begin{tabular}{cc}
      \begin{minipage}[t]{0.66\hsize}
        \centering
        \includegraphics[width=8.5cm,pagebox=cropbox,clip]{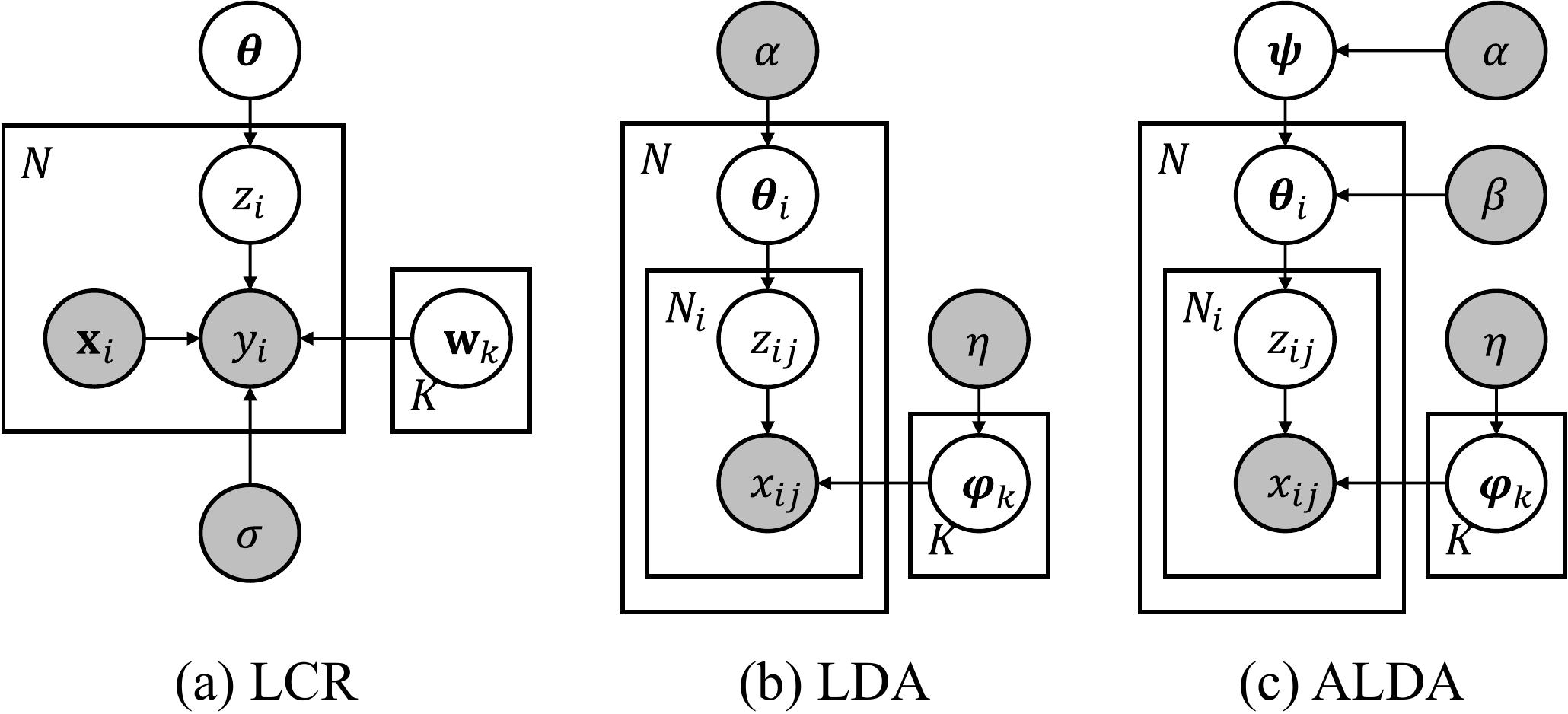}
        \caption{Mixture Models}
        \label{fig:mm}
      \end{minipage}
      \hfill
      \begin{minipage}[t]{0.3\hsize}
        \centering
        \includegraphics[width=3.3cm,pagebox=cropbox,clip]{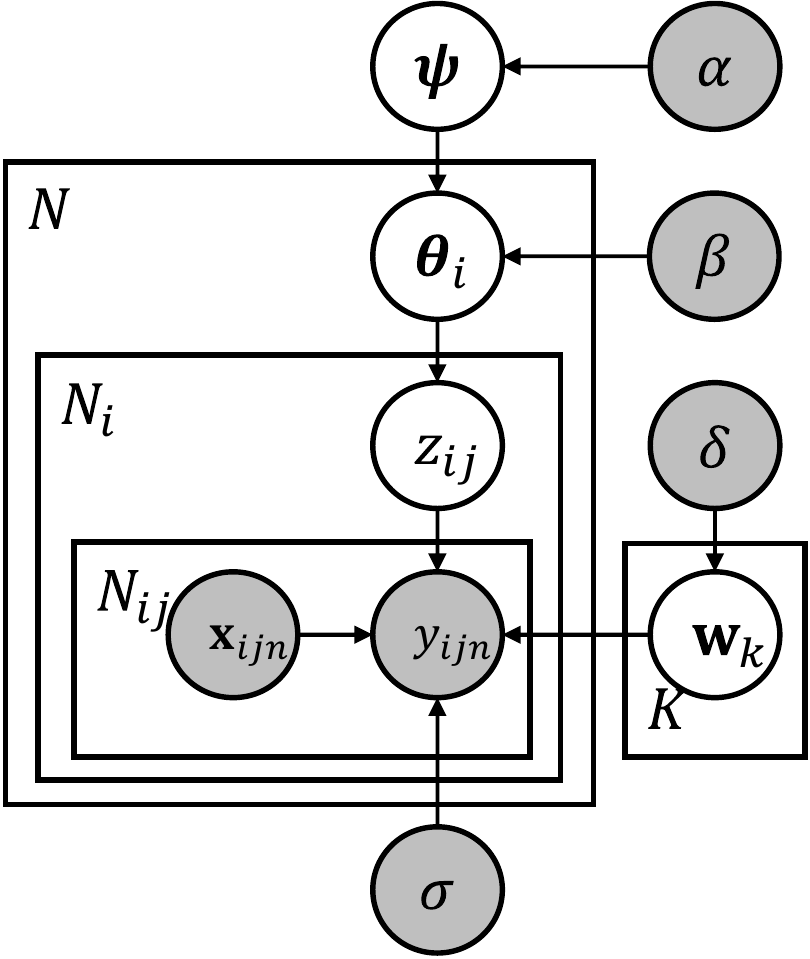}
        \caption{Hierarchical Latent Class Regression}
        \label{fig:HLCR}
      \end{minipage}
    \end{tabular}
\end{figure}

{\bf Latent Class Regression (LCR)} Latent Class Regression \cite{wedel1994review,Grun2008} (Figure \ref{fig:mm}(a)) is a supervised learning model that can be defined as a mixture of regression models $\Sigma_{k=1}^K \theta_k p(y|\mathbf{x}, \mathbf{w}_k)$ $(k \in [K])$, in which $\mathbf{w}_k$ is the parameter ($\boldsymbol{\varphi}_k$) of the $k$-th model. In different unsupervised learning techniques, like the GMM, each model is a probability on a target $y$ given an observation $\mathbf{x}$ where the model could be any regression function. If, for instance, $p(y|\mathbf{x}, \mathbf{w}_k)$ is a linear regression model, the distribution of $y$ can be denoted as $\Sigma_{k=1}^K \theta_k \mathcal{N}(\mathbf{w}_k^\top \cdot \mathbf{x}, \sigma^2)$, where $\mathbf{w}_k$ is the coefficient of the $k$-th linear model and $\sigma^2$ is the variance of the white noise. As with the GMM, LCR can be trained by the EM algorithm.

{\bf Latent Dirichlet Allocation (LDA)} One of the most famous topic models is Latent Dirichlet Allocation (LDA) \cite{blei2003latent} (Figure \ref{fig:mm}(b)). A topic model generally handles a set of documents, each of which is composed of a set of words from a dictionary (the set of all possible words). This kind of document-word data can be regarded as a hierarchical structure with two layers. In LDA, each topic $k \in [K]$ is defined as a Categorical distribution over words in a dictionary with probability  $\boldsymbol{\varphi}_k$.
Although EM cannot be directly used for inferring LDA, various inference techniques have been proposed, such as Variational Bayesian (VB) inference, Collapsed Variational Bayesian (CVB) inference, and Collapsed Gibbs Sampling (CGS).
LDA with an asymmetric Dirichlet prior (ALDA) \cite{wallach2009rethinking} (Figure \ref{fig:mm}(c)) is similar to general LDA with a symmetric prior.
ALDA assumes documents are generated in four steps: 1) an asymmetric Categorical distribution (prior) $\boldsymbol{\psi}$ is sampled from a Dirichlet distribution with a symmetric parameter $\alpha(\frac{1}{K}, \cdots, \frac{1}{K})$; 2) for document $i$, a Categorical distribution $\boldsymbol{\theta}_i$ is sampled from a Dirichlet distribution with a parameter $\beta \boldsymbol{\psi}$; 3) for word $j$ in document $i$, a latent cluster label $z_{ij}$ is sampled from $\boldsymbol{\theta}_i$; and 4) the word $j$ in document $i$, denoted as $x_{ij}$, is then sampled from the $z_{ij}$-th topic, $\boldsymbol{\varphi}_{z_{ij}}$.
$\boldsymbol{\psi}$ is the prior of all documents, so it can be regarded as a global distribution of topics.
ALDA increases the robustness of topic models and can be efficiently inferred by Collapsed Gibbs Sampling.

\section{Hierarchical Latent Class Regression}\label{sec.HLCR}
Our goal is to propose a regression model for predicting data with a hierarchical structure. 
The hierarchical structure considered in this paper is similar to, but more general than, the document-word structure in the topic model. In the Wi-Fi connection example, smartphones (agent) and Wi-Fi access points (entity) are similar to documents and words in topic models, respectively. In other words, the data from each smartphone contains multiple Wi-Fi access points; however, unlike in the document-word structures in which each word is a simple record, each smartphone usually accesses each Wi-Fi access point more than once. We call each connection an event. In this paper, we generally call this three-layer hierarchical structure an agent-entity-event structure. Another difference is that our problem is a regression problem, so each event is composed of a feature vector, denoting the condition of the connection, and a target value, denoting the quality of the event.

\subsection{Model}
In order to handle the regression problem on the agent-entity-event hierarchical data, we propose a Hierarchical Latent Class Regression (HLCR) model by introducing the mechanism of hierarchical structure in ALDA to LCR. It is assumed that there are $N$ agents, each of which is labeled by $i \in [N]$. Each agent $i$ contains $N_i$ entities, each of which is labeled by $j \in [N_i]$ and contains $N_{ij}$ events. Moreover, each of the $N_{ij}$ events is composed of an $F$-dimensional column vector $\mathbf{x}_{ijn}$ and a target scalar $y_{ijn}$ ($n \in N_{ij}$).
In the mixture models introduced in the previous section, each record corresponds to a latent variable $z_{*}$, denoting its cluster. In our HLCR, however, we assume that all events corresponding to the same agent-entity pair belong to the same cluster, since they are assumed to have similar behavior. More specifically, all $\mathbf{x}_{ijn}$s and $y_{ijn}$s ($\forall n \in [N_{ij}]$) for any particular $i$ and $j$ share the same cluster label $z_{ij}$.
We denote the set of all $\mathbf{x}_{ijn}$ and the set of all $y_{ijn}$ ($n \in [N_{ij}]$) as an $(N_{ij} \times F)$-matrix $\mathbf{X}_{ij}$, each row of which is $\mathbf{x}_{ijn}^\top$, and $N_{ij}$-dimensional column vector $\mathbf{y}_{ij}$.

The graphical model of HLCR (Figure \ref{fig:HLCR}) assumes that data are generated in the following steps:
1) each topic $k \in [K]$ samples an $F$-dimensional vector $\mathbf{w}_k$ $(k \in [K])$ from a normal distribution, $\mathbf{w}_k \sim \mathcal{N}(0, \delta^2 \mathbf{I})$;
2) the global Categorical distribution $\boldsymbol{\psi}$ is sampled from a Dirichlet distribution with a symmetric parameter, $\boldsymbol{\psi} \sim \mbox{Dir}\left( \alpha(\frac{1}{K}, \cdots, \frac{1}{K}) \right)$;
3) for agent $i \in [N]$, a Categorical distribution $\boldsymbol{\theta}_i$ is sampled from a Dirichlet distribution, $\boldsymbol{\theta}_i \sim \mbox{Dir}(\beta \boldsymbol{\psi})$;
4) for entity $j \in [N_i]$ in agent $i$, a latent cluster label $z_{ij}$ is sampled from a Categorical distribution, $z_{ij} \sim \mbox{Categorical}(\boldsymbol{\theta}_i)$;
and 5) for event $n \in [N_{ij}]$ in entity $j$ in agent $i$, the target value $y_{ijn}$ is then generated by adding a Gaussian noise to $\mathbf{w}_{z_{ij}}^\top \mathbf{x}_{ijn}$ corresponding to the $z_{ij}$-th topic, $y_{ijn} \sim \mathcal{N} ( \mathbf{w}_{z_{ij}}^\top \mathbf{x}_{ijn}, \sigma^2 )$.
The prior distribution of $z_{ij}$ in HLCR is the same as that in ALDA, and the probability of $y_{ijn}$ in HLCR is similar to that in LCR except for two differences. First, $N_{ij}$ events $\mathbf{y}_{ij}$ in HLCR share one $z_{ij}$, while one word, $y_{ij}$, in LCR owns one $z_{ij}$. Second, we add a prior $\delta$ for $\mathbf{w}_k$ in HLCR for deriving the Collapsed Gibbs Sampling algorithm, while $\mathbf{w}_k$ in LCR has no prior since it is not necessary for an EM algorithm.

\subsection{Inference}
In each iteration of Collapsed Gibbs Sampling, a new cluster label $z_{ij}^{(new)}$ for every particular $i \in [N]$ and $j \in [N_i]$ is sampled sequentially, with all other cluster labels, denoted as $\mathbf{z}_{\setminus ij} (= \mathbf{z} \setminus \{ z_{ij} \})$, being fixed.
Hence, we need to evaluate the conditional probability $p(z_{ij}^{(new)}=k | \mathbf{z}_{\setminus ij}, \mathbf{X}, \mathbf{y}, \alpha, \beta, \delta, \sigma)$, in which all other random variables, $\boldsymbol{\psi}$, $\boldsymbol{\theta}_i$ and $\mathbf{w}_k$, are integrated out. From Bayes' theorem, we have
\begin{equation}\label{eq.gibbs}
p(z_{ij}^{(new)}=k | \mathbf{z}_{\setminus ij}, \mathbf{X}, \mathbf{y}, \alpha, \beta, \delta, \sigma) \propto p(z_{ij}^{(new)}=k | \mathbf{z}_{\setminus ij}, \alpha, \beta) p(\mathbf{y}_{ij} | \mathbf{z}^{(new)}, \mathbf{X}, \mathbf{y}_{\setminus ij}, \delta, \sigma),
\end{equation}
where $\mathbf{z}^{(new)}$ denotes $\mathbf{z}_{\setminus ij} \cup \{ z_{ij}^{(new)} \}$.

The first part in (\ref{eq.gibbs}) $p(z_{ij}^{(new)} | \mathbf{z}_{\setminus ij}, \alpha, \beta)$ can be computed by integrating out $\boldsymbol{\psi}$ and $\boldsymbol{\theta}_i$ from $p(z_{ij}^{(new)}, \boldsymbol{\psi}, \boldsymbol{\theta}_i | \mathbf{z}_{\setminus ij}, \alpha, \beta)$, which is equal to $p(\boldsymbol{\psi} | \alpha) p(\boldsymbol{\theta}_i | \beta, \boldsymbol{\psi}) p(z_{ij} | \boldsymbol{\theta}_i)$. Since the generative process of $z_{ij}^{(new)}$ is completely the same as that in \cite{wallach2009rethinking}, we obtain the same result as follows.
\begin{equation}\label{eq.prior}
p(z_{ij}^{(new)}=k | \mathbf{z}_{\setminus ij}, \alpha, \beta) = \frac{N_i^{(k)} + \beta \frac{N^{(k)}+\frac{\alpha}{K}}{\Sigma_k N^{(k)} + \alpha}}{\Sigma_k N_i^{(k)} + \beta},
\end{equation}
where $N_i^{(k)}$ denotes the number of entities in agent $i$, whose cluster labels $z_{ij}$ are equal to $k$, i.e., $N_i^{(k)} = \sum_{j \in [N_i]}^{z_{ij}=k} 1$, and $N^{(k)}$ denotes that number in all agents, i.e., $N^{(k)} = \sum_{i \in [N]} N_i^{(k)}$.

The second part $p(\mathbf{y}_{ij} | \mathbf{z}^{(new)}, \mathbf{X}, \mathbf{y}_{\setminus ij}, \delta, \sigma)$, on the other hand, is equal to $p(\mathbf{y}_{ij} | z_{ij}^{(new)}=k, \mathbf{X}_{ij}, \mathbf{X}_{\setminus ij}^{(k)}, \mathbf{y}_{\setminus ij}^{(k)}, \delta, \sigma)$. The following theorem can be used to compute this probability.

\newtheorem{theorem}{Theorem}
\theoremstyle{empty}
\newtheorem{refproof}{Proof}

\begin{theorem}\label{thm.likelihood}
For any particular $i \in [N]$ and $j \in [N_i]$, let $\mathbf{X}_{ij}$ and $\mathbf{y}_{ij}$ be the data corresponding to $i$ and $j$. For $k \in [K]$, let $\mathbf{X}_{\setminus ij}^{(k)}$ and $\mathbf{y}_{\setminus ij}^{(k)}$ be the data whose cluster labels are equal to $k$ except $\mathbf{X}_{ij}$ and $\mathbf{y}_{ij}$. Then, the conditional probability of $\mathbf{y}_{ij}$ given the new cluster label $z_{ij}^{(new)}$ is
\begin{equation}
p \left( \mathbf{y}_{ij} | z_{ij}^{(new)}=k, \mathbf{X}_{ij}, \mathbf{X}_{\setminus ij}^{(k)}, \mathbf{y}_{\setminus ij}^{(k)}, \delta, \sigma \right) = \prod_{n=1}^{N_{ij}} p_{ijn}(y = y_{ijn}),
\end{equation}
where $p_{ijn}(y)$ obeys a normal distribution as the following.
\begin{eqnarray}
p_{ijn}(y)& \sim & \mathcal{N} \left( \mathbf{b}_{n-1}^\top \mathbf{A}_{n-1}^{-1} \mathbf{x}_{ijn}, \sigma^2 + \frac{1}{\sigma^2} \mathbf{x}_{ijn}^\top \mathbf{A}_{n-1}^{-1} \mathbf{x}_{ijn} \right), \label{eq:likelihood} \\
\mathbf{A}_n & = & \frac{1}{\delta^2} \mathbf{I} + \frac{1}{\sigma^2} \left( \mathbf{X}_{\setminus ij}^{(k)\top} \mathbf{X}_{\setminus ij}^{(k)} + \sum_{m=1}^n \mathbf{x}_{ijm} \mathbf{x}_{ijm}^\top \right) \label{eq.A_n}, \\
\mathbf{b}_n & = & \frac{1}{\sigma^2} \left( \mathbf{X}_{\setminus ij}^{(k)\top} \mathbf{y}_{\setminus ij}^{(k)} + \sum_{m=1}^n \mathbf{x}_{ijm} y_{ijm} \right).
\end{eqnarray}
\end{theorem}

It is therefore straightforward that
\begin{eqnarray}
\mathbf{A}_n = \mathbf{A}_{n-1} + \frac{1}{\sigma^2} \mathbf{x}_{ijn} \mathbf{x}_{ijn}^\top &, &\mathbf{b}_n = \mathbf{b}_{n-1} + \frac{1}{\sigma^2} \mathbf{x}_{ijn} y_{ijn}. \label{eq.recursive.Ab}
\end{eqnarray}

Furthermore, in order to evaluate this normal distribution, we have to compute the inverse of ($F \times F$)-matrix $\mathbf{A}_n$, where the computational cost is generally $O(F^3)$. By combining the Woodbury matrix identity with (\ref{eq.recursive.Ab}), we obtain the following recursive rules that reduce the computational cost to $O(F^2)$.
\begin{eqnarray}
\mathbf{A}_n^{-1} & = & \mathbf{A}_{n-1}^{-1} - \frac{1}{\sigma^2} \mathbf{A}_{n-1}^{-1} \mathbf{x}_{ijn} (1 + \frac{1}{\sigma^2} \mathbf{x}_{ijn}^\top \mathbf{A}_{n-1}^{-1} \mathbf{x}_{ijn})^{-1} \mathbf{x}_{ijn}^\top \mathbf{A}_{n-1}^{-1}, \label{eq:recuisive.2} \\
\mathbf{A}_{n-1}^{-1} & = & \mathbf{A}_n^{-1} + \frac{1}{\sigma^2} \mathbf{A}_n^{-1} \mathbf{x}_{ijn} (1 - \frac{1}{\sigma^2} \mathbf{x}_{ijn}^\top \mathbf{A}_n^{-1} \mathbf{x}_{ijn})^{-1} \mathbf{x}_{ijn}^\top \mathbf{A}_n^{-1}. \label{eq:recuisive.1}
\end{eqnarray}


Let $\mathbf{D}^{(k)} = \frac{1}{\delta^2} \mathbf{I} + \frac{1}{\sigma^2} \mathbf{X}^{(k)\top} \mathbf{X}^{(k)}$, $\mathbf{c}^{(k)} = \frac{1}{\sigma^2} \mathbf{X}^{(k)\top} \mathbf{y}^{(k)}$ and $\mathbf{H}^{(k)} = {\mathbf{D}^{(k)}}^{-1}$, which can be computed from the data whose current labels are equal to $k$. Hence, for any particular $i$ and $j$, there are two cases:
1) if the current $z_{ij} \neq k$, $\mathbf{D}^{(k)}$ and $\mathbf{c}^{(k)}$ do NOT contain $\mathbf{X}_{ij}$ and $\mathbf{y}_{ij}$, then $\mathbf{A}_0^{-1} = \mathbf{H}^{(k)}$ and $\mathbf{b}_0 = \mathbf{c}^{(k)}$;
and 2) if the current $z_{ij} = k$, $\mathbf{D}^{(k)}$ and $\mathbf{c}^{(k)}$ contain $\mathbf{X}_{ij}$ and $\mathbf{y}_{ij}$, then $\mathbf{A}_{N_{ij}}^{-1} = \mathbf{H}^{(k)}$ and $\mathbf{b}_{N_{ij}} = \mathbf{c}^{(k)}$. Then, one can recursively compute $\mathbf{A}_0^{-1}$ and $\mathbf{b}_0$ with (\ref{eq:recuisive.1}) and (\ref{eq.recursive.Ab}).
By obtaining $\mathbf{A}_0^{-1}$ and $\mathbf{b}_0$, we can recursively compute $\mathbf{A}_n^{-1}$ and $\mathbf{b}_n$ ($\forall n \in [N_{ij}]$) with (\ref{eq:recuisive.2}) and (\ref{eq.recursive.Ab}), and by substituting them into (\ref{eq:likelihood}), we can compute the mean and variance of the normal distribution, and then evaluate the conditional probability of $y_{ijn}$ (\ref{eq:likelihood}). Substituting this result and the prior of $z_{ij}^{(new)}$ (\ref{eq.prior}) into (\ref{eq.gibbs}), we can compute the conditional probability of $z_{ij}^{(new)}=k$ ($k \in [K]$). Algorithm \ref{alg:LabelSampling} shows how to sample the new label $z_{ij}^{(new)}$ for specific $i$ and $j$.

\begin{algorithm}
\caption{Label Sampling}
\label{alg:LabelSampling}
\begin{algorithmic}[1]
 \renewcommand{\algorithmicrequire}{\textbf{Input:}}
 \renewcommand{\algorithmicensure}{\textbf{Output:}}
 \Require{$\mathbf{X}_{ij}$: $(N_{ij} \times F)$-matrix, $\mathbf{y}_{ij}$: $N_{ij}$-vector, $\mathbf{H} = \{ \mathbf{H}^{(1)}, \dots, \mathbf{H}^{(K)} \}$: a set of $(F \times F)$-matrices, $\mathbf{c} = \{ \mathbf{c}^{(1)}, \cdots, \mathbf{c}^{(K)} \}$: a set of $F$-vectors, $\alpha, \beta, \delta, \sigma > 0$, $K \geq 0$}
 \Ensure{$z_{ij}^{(new)}$: sampled new label}
 \Function{SampleLabel}{$\mathbf{X}_{ij}, \mathbf{y}_{ij}, \mathbf{H}, \mathbf{c}, \alpha, \beta, \delta, \sigma, K$}
  \For{$k = 1$ to $K$}
   \State compute $p_{ij}^{(k)} = p(z_{ij}^{(new)}=k | \mathbf{z}_{\setminus ij}, \alpha, \beta)$\Comment{Refer to (\ref{eq.prior})}
   \State $\mathbf{A}_0^{-1} \gets \mathbf{H}^{(k)}$, $\mathbf{b}_0 \gets \mathbf{c}^{(k)}$
   \For{$n = 1$ to $N_{ij}$}
    \State $\mathbf{b}_n \gets \mathbf{b}_{n-1} + \frac{1}{\sigma^2} \mathbf{x}_{ijn} y_{ijn}$
    \State $\mathbf{A}_n^{-1} \gets \mathbf{A}_{n-1}^{-1} - \frac{1}{\sigma^2} \mathbf{A}_{n-1}^{-1} \mathbf{x}_{ijn} (1 + \frac{1}{\sigma^2} \mathbf{x}_{ijn}^\top \mathbf{A}_{n-1}^{-1} \mathbf{x}_{ijn})^{-1} \mathbf{x}_{ijn}^\top \mathbf{A}_{n-1}^{-1}$\Comment{Refer to (\ref{eq:recuisive.2})}
    \State compute $p_{ijn}$\Comment{Refer to (\ref{eq:likelihood})}
    \State $p_{ij}^{(k)} \gets p_{ij}^{(k)} \cdot p_{ijn}$
   \EndFor
  \EndFor
  \State \Return $z_{ij}^{(new)} \sim \textrm{Categorical} (p_{ij}^{(1)}, p_{ij}^{(2)}, \cdots, p_{ij}^{(K)})$
 \EndFunction
\end{algorithmic}
\end{algorithm}

We compute matrices $\mathbf{H}$ and vectors $\mathbf{c}$ and pass the result to Algorithm \ref{alg:LabelSampling}. For each $k \in [K]$, $p_{ij}^{(k)}$ is initialized to the prior of $z_{ij}^{(new)}$ in Line 3. $\mathbf{b}_n$ and $\mathbf{A}_n$ are updated in Lines 6-7, then the probability of $y_{ijn} (n \in [N_{ij}])$ is computed in Line 9. Next, $p_{ij}^{(k)}$ becomes the conditional probability of $z_{ij}^{(new)}$ (\ref{eq.gibbs}). Finally, a $z_{ij}^{(new)}$ is sampled from a categorical distribution $(p_{ij}^{(1)}, p_{ij}^{(2)}, \cdots, p_{ij}^{(K)})$ and is returned.

\subsection{Training Algorithm}
\label{sec.algorithm}
Using Algorithm \ref{alg:LabelSampling}, we can easily implement the training algorithm of HLCR in Algorithm \ref{alg:HLCR}. The algorithm is started by randomly initializing all latent variables $z_{ij}$s ($\forall i \in [N]$ and $\forall j \in [N_i]$) in Lines 1-3, then $\mathbf{c}$ and $\mathbf{H}$ are initialized based on these latent variables in Lines 5-6. In each iteration $t \in [T]$, for each agent $i \in [N]$ and each entity $j \in [N_i]$, we remove $\mathbf{X}_{ij}$ and $\mathbf{y}_{ij}$ from $\mathbf{c}^{(z_{ij})}$ and $\mathbf{H}^{(z_{ij})}$ corresponding to the current $z_{ij}$ in Lines 10-12, then sample a new latent variables $z_{ij}$ with function SampleLabel in Algorithm \ref{alg:LabelSampling} in Line 13, and finally add $\mathbf{X}_{ij}$ and $\mathbf{y}_{ij}$ to $\mathbf{c}^{(z_{ij})}$ and $\mathbf{H}^{(z_{ij})}$ corresponding to this new $z_{ij}$ in Lines 14-16.
\begin{algorithm}
\caption{Hierarchical Latent Class Regression (HLCR)}
\label{alg:HLCR}
\begin{algorithmic}[1]
 \renewcommand{\algorithmicrequire}{\textbf{Input:}}
 \Require{$\mathbf{X}$: $(\sum_{i,j} N_{ij} \times F)$-matrix, $\mathbf{y}$: $(\sum_{i,j} N_{ij})$-vector, $\alpha, \beta, \delta, \sigma > 0$, $K \geq 0$}
 \Procedure{HLCR}{$\mathbf{X}$, $\mathbf{y}$, $\alpha$, $\beta$, $\delta$, $\sigma$, $K$}
  \For{$i = 1$ to $N$}
   \For{$j = 1$ to $N_i$}
    \State randomly initialize $z_{ij} \sim \mathcal{U}(1, K)$
   \EndFor
  \EndFor
  \For{$k = 1$ to $K$}
   \State $\mathbf{c}^{(k)} \gets \frac{1}{\sigma^2} \mathbf{X}^{(k)\top} \mathbf{y}^{(k)}$, $\mathbf{H}^{(k)} \gets (\frac{1}{\delta^2} \mathbf{I} + \frac{1}{\sigma^2} \mathbf{X}^{(k)\top} \mathbf{X}^{(k)})^{-1}$
  \EndFor
  \For{$t = 1$ to $T$}
   \For{$i = 1$ to $N$}
    \For{$j = 1$ to $N_i$}
     \For{$n = N_{ij}$ to $1$}\Comment{Remove data $i,j$ from current cluster $z_{ij}$}
      \State $\mathbf{H}^{(z_{ij})} \gets \mathbf{H}^{(z_{ij})} + \frac{1}{\sigma^2} \mathbf{H}^{(z_{ij})} \mathbf{x}_{ijn} (1 - \frac{1}{\sigma^2} \mathbf{x}_{ijn}^\top \mathbf{H}^{(z_{ij})} \mathbf{x}_{ijn})^{-1} \mathbf{x}_{ijn}^\top \mathbf{H}^{(z_{ij})}.$\Comment{(\ref{eq:recuisive.1})}
      \State $\mathbf{c}^{(z_{ij})} \gets \mathbf{c}^{(z_{ij})} - \frac{1}{\sigma^2} \mathbf{x}_{ijn} y_{ijn}$
     \EndFor
     \State $z_{ij} \gets$ \Call{SampleLabel}{$\mathbf{X}_{ij}$, $\mathbf{y}_{ij}$, $\mathbf{H}$, $\mathbf{c}$, $\alpha$, $\beta$, $\delta$, $\sigma$, $K$} \Comment{Algorithm \ref{alg:LabelSampling}}
     \For{$n = 1$ to $N_{ij}$}\Comment{Add data $i,j$ to new cluster $z_{ij}$}
      \State $\mathbf{H}^{(z_{ij})} \gets \mathbf{H}^{(z_{ij})} - \frac{1}{\sigma^2} \mathbf{H}^{(z_{ij})} \mathbf{x}_{ijn} (1 + \frac{1}{\sigma^2} \mathbf{x}_{ijn}^\top \mathbf{H}^{(z_{ij})} \mathbf{x}_{ijn})^{-1} \mathbf{x}_{ijn}^\top \mathbf{H}^{(z_{ij})}.$\Comment{(\ref{eq:recuisive.2})}
      \State $\mathbf{c}^{(z_{ij})} \gets \mathbf{c}^{(z_{ij})} + \frac{1}{\sigma^2} \mathbf{x}_{ijn} y_{ijn}$
     \EndFor
    \EndFor
   \EndFor
  \EndFor
 \EndProcedure
\end{algorithmic}
\end{algorithm}

\subsection{Prediction}
After training the data with Algorithm \ref{alg:HLCR}, we obtain cluster label $z_{ij}$s for all agents $i \in [N]$ and entities $j \in [N_i]$, $\mathbf{D}^{(k)}$, $\mathbf{c}^{(k)}$ and $\mathbf{H}^{(k)}$ for all $k \in [K]$.
Suppose there are $N_{ij}$ data for a particular agent $i$ and entity $j$. If we get a new feature $\mathbf{x}_{ij(N_{ij}+1)}$, its target $y_{ij(N_{ij}+1)}$ can be predicted with HLCR from the variables obtained from the algorithm. HLCR predicts the target value in two steps. 1) Select a proper cluster label for $\mathbf{x}_{ij(N_{ij}+1)}$ based on previous data $\mathbf{X}_{ij}$ and $\mathbf{y}_{ij}$ with Algorithm \ref{alg:LabelSampling}. In fact, if there have been previous data of $\mathbf{X}_{ij}$ and $\mathbf{y}_{ij}$ in the training set, we can directly use the corresponding latent variable $z_{ij}$ without sampling it again. Then 2) using the variables corresponding to $z_{ij}$, i.e., $\mathbf{D}^{(z_{ij})}$, $\mathbf{c}^{(z_{ij})}$ and $\mathbf{H}^{(z_{ij})}$, we can predict the target value as follows.

\begin{theorem}\label{thm.prediction}
For any particular $i \in [N]$ and $j \in [N_i]$, let $\mathbf{X}$ and $\mathbf{y}$ be all data, $z_{ij}$ be the cluster label corresponding to $i$ and $j$, and $\mathbf{D}^{(k)}$, $\mathbf{c}^{(k)}$ and $\mathbf{H}^{(k)}$ $(k \in [K])$ be the variables trained by Algorithm \ref{alg:HLCR}. Then, given a new feature $\mathbf{x}_{ij(N_{ij}+1)}$, the conditional expected value of the target $y_{ij(N_{ij}+1)}$ is
\begin{equation}
\hat{y}_{ij(N_{ij}+1)} = E(y_{ij(N_{ij}+1)} | \mathbf{x}_{ij(N_{ij}+1)}, z_{ij}, \mathbf{X}, \mathbf{y}, \delta, \sigma) = {\mathbf{c}^{(z_{ij})}}^\top \mathbf{H}^{(z_{ij})} \mathbf{x}_{ij(N_{ij}+1)}.
\end{equation}
\end{theorem}
This theorem indicates that the prediction result is the same as the solution of the Ridge regression.

\paragraph{Discussion}
HLCR is a mixture of linear regressions.
Since HLCR is more expressive, it can be used to efficiently predict noisy data with various structures. The mechanism of HLCR is different from that of general regression models, such as Deep Neural Networks (DNN). A regression model is generally a regression function that maps a feature to a target, and hence, given the same feature, the predicted target value will be unique. However, in HLCR, different predictions can be made for identical input features if they originate from agent-entities that belong to different clusters (models). This property makes our HLCR fit hierarchical data better than general regression models do.



\begin{algorithm}
\caption{Federated Hierarchical Latent Class Regression (FedHLCR)}
\label{alg:FedHLCR}
\begin{algorithmic}[1]
 \renewcommand{\algorithmicrequire}{\textbf{Input:}}
 \Require{$\mathbf{X}$: $(\sum_{i,j} N_{ij} \times F)$-matrix, $\mathbf{y}$: $(\sum_{i,j} N_{ij})$-vector, $\alpha, \beta, \delta, \sigma > 0$, $K \geq 0$, $0 \leq \gamma \leq 1$: learning rate}
 \Procedure{FedHLCR}{$\mathbf{X}, \mathbf{y}, \alpha, \beta, \delta, \sigma, K, \gamma$}
  \State \underline{server} initialize $\mathbf{H}_{(0)} \gets \{ \delta^2 \mathbf{I}, \delta^2 \mathbf{I}, \cdots, \delta^2 \mathbf{I} \}$ and $\mathbf{c}_{(0)} \gets \{ \mathbf{0}, \mathbf{0}, \cdots, \mathbf{0} \}$ $(t=0)$
  \For{$t = 1$ to $T$}
   \State \underline{server} initialize $\mathbf{D} \gets \{ \frac{1}{\delta^2} \mathbf{I}, \frac{1}{\delta^2} \mathbf{I}, \cdots, \frac{1}{\delta^2} \mathbf{I} \}$ and $\mathbf{c} \gets \{ \mathbf{0}, \mathbf{0}, \cdots, \mathbf{0} \}$
   \For{\underline{agent} $i \in [N]$ in parallel}
    \State receive $\mathbf{H}_{(t-1)}$ and $\mathbf{c}_{(t-1)}$ from \underline{server}
    \State initialize $\mathbf{D}_i \gets \mathbf{0}$ and $\mathbf{c}_i \gets \mathbf{0}$
    \For{$j = 1$ to $N_i$}
     \State $z_{ij} \gets$ \Call{SampleLabel}{$\mathbf{X}_{ij}$, $\mathbf{y}_{ij}$, $\mathbf{H}_{(t-1)}$, $\mathbf{c}_{(t-1)}$, $\alpha$, $\beta$, $\delta$, $\sigma$, $K$} \Comment{Algorithm \ref{alg:LabelSampling}}
     \State $\mathbf{c}_i^{(z_{ij})} \gets \mathbf{c}_i^{(z_{ij})} + \frac{1}{\sigma^2} \mathbf{x}_{ijn} y_{ijn}$, $\mathbf{D}_i^{(z_{ij})} \gets \mathbf{D}_i^{(z_{ij})} + \frac{1}{\sigma^2} \mathbf{x}_{ijn} \mathbf{x}_{ijn}^\top$
    \EndFor
    \State send $\mathbf{c}_i$ and $\mathbf{D}_i$ to \underline{server}
    \State \underline{server} $\mathbf{c} \gets \mathbf{c} + \mathbf{c}_i$, $\mathbf{D} \gets \mathbf{D} + \mathbf{D}_i$
   \EndFor
   \If{$t = 1$}
    \State \underline{server} $\mathbf{c}_{(1)} \gets \mathbf{c}$, $\mathbf{D}_{(1)} \gets \mathbf{D}$
   \Else
    \State \underline{server} $\mathbf{c}_{(t)} \gets (1-\gamma) \mathbf{c}_{(t-1)} + \gamma \mathbf{c}$, $\mathbf{D}_{(t)} \gets (1-\gamma) \mathbf{D}_{(t-1)} + \gamma \mathbf{D}$
   \EndIf
   \State $\mathbf{H}_{(t)}^{(k)} = {\mathbf{D}_{(t)}^{(k)}}^{-1}$, for $\forall k \in [K]$
  \EndFor
 \EndProcedure
\end{algorithmic}
\end{algorithm}

\section{Federated Hierarchical Latent Class Regression}\label{sec.FedHLCR}
Federated Hierarchical Latent Class Regression (FedHLCR) is shown in Algorithm \ref{alg:FedHLCR}. In the beginning, the server initializes $\mathbf{c}_{(0)}$ and $\mathbf{H}_{(0)}$ without any data in Line 2. In the beginning of each iteration,
each agent $i$ receives the global model $\mathbf{H}_{(t-1)}$ and $\mathbf{c}_{(t-1)}$ trained in previous iteration from the server (Line 6), and trains its local data in parallel based on the global model (Lines 7-10), then sends the training results to the server (Line 11), and finally, the server accumulates the local training results received from agents to the global model, $\mathbf{D}$ and $\mathbf{c}$, at the end of the iteration (Line 12). In Federated Learning, not all agents participate the training process in each iteration and no one can guarantee that the agent set in each iteration does not change (Line 5). In order to ensure the convergence of the algorithm, we smoothly update the model by using a learning rate $\gamma$ in Line 16. 

There are several differences between FedHLCR in Algorithm \ref{alg:FedHLCR} and centralized HLCR in Algorithm \ref{alg:HLCR}. First, HLCR updates intermediate data $\mathbf{H}$ and $\mathbf{c}$ whenever $z_{ij}$ is sampled, while each agent $i$ in FedHLCR trains its data $\mathbf{X}_{ij}$ and $\mathbf{y}_{ij}$ ($\forall j \in [N_{i}]$) independently and the server only updates $\mathbf{H}$ and $\mathbf{c}$ once in each iteration. This process saves communication costs and efficiently protects privacy. Second, the data for each entity in each agent in HLCR are trained once in each iteration, and $\mathbf{H}$ and $\mathbf{c}$ are updated in summation, while only a part of agents in FedHLCR participate the training process and the server updates the model in a smooth way (Line 16). This approach makes FedHLCR smoothly converge. Finally, HLCR removes $\mathbf{X}_{ij}$ and $\mathbf{y}_{ij}$ from $\mathbf{H}$ and $\mathbf{c}$ before sampling $z_{ij}$ for $\forall i \in [N]$ and $j \in [N_i]$, while FedHLCR does not remove them before sampling since the weight for each record decreases after several update steps (Line 16), thus this simplifies our algorithm.

%% file: experiment.tex
We use synthetic and real data in order to systematically test the \textsc{FedHLCR} algorithm performance on heterogeneous datasets.
All experiments are performed on a commodity machine with two Intel\textregistered Xeon\textregistered CPUs E5-2690 v3 @ 2.60GHz and 64GB of memory. 


\paragraph{Sysnthetic data}

\begin{figure*}\center
    \caption{
    Accuracy (MSE) of \textsc{FedHLCR} on the \textsc{SynthHLCR} dataset.
    (a) MSE w.r.t. different data size and noise.
    (b) MSE with 15\% of agents sampled at each iteration.
    (c) MSE w.r.t. different $\gamma$.
    \label{fig:synthlcr}}
    \center
    \includegraphics[width=\textwidth]{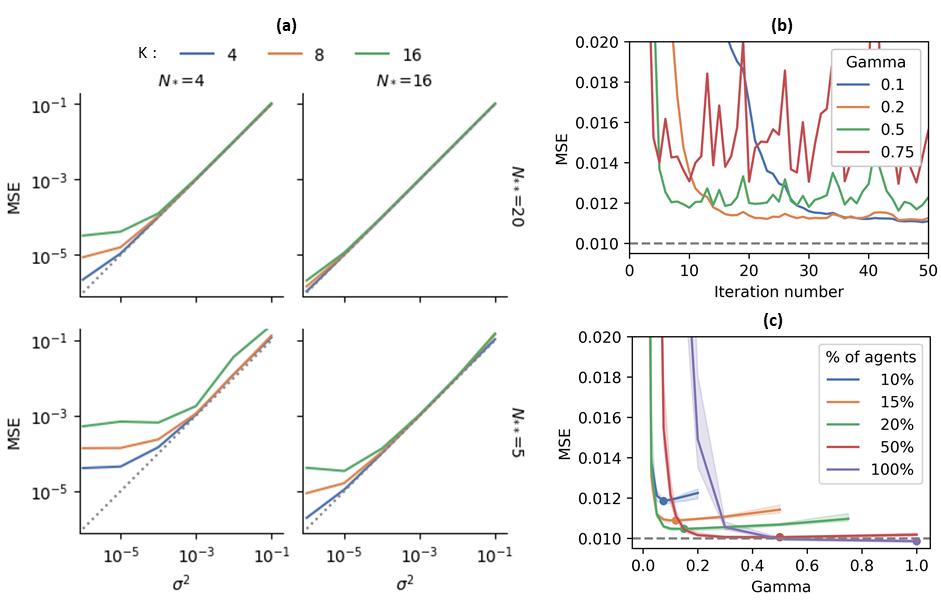}
\end{figure*}

The synthetic dataset \textsc{SynthHLCR} is generated according to the model in Figure~\ref{fig:HLCR}.
We control the average number of entities per agent by $N_*$ and the average number of events per entity by $N_{**}$.
We generate a series of such datasets with different noise level $\sigma$, cluster number $K$, and average category number per agent $N_{*}$, and the dataset is generated using 128 agents and 128 entities for various choices of $N_*$ and $N_{**}$.
We then perform a 5-fold cross-validation on the generated data and train the \textsc{FedHLCR} algorithm with the same parameter choice as for the generation process.
We observe that the \textsc{FedHLCR} converges in less than 10 iterations. The results reported in Figure~\ref{fig:synthlcr} (a) show that the converged model at high noise is optimal when agents have 20 datapoints on average.
For $N_*=4$, $N_{**}=5$, $K=16$, and high $\sigma$, the dataset has a signal-to-noise ratio too low to correctly characterize the model, leading to agent-entity misclustering.
When $\sigma$ is small, we observe a deviation to the optimal solution which is caused by relatively big prior $\delta$. 

\paragraph{Federated simulation}

One aspect of Federated Learning is the agent failure to report on the one hand, and the deliberate agent sampling by the orchestrator on the other. 
We simulate this by randomly sampling a certain ratio of agents among all agents in each iteration.
In Figure \ref{fig:synthlcr} (b) and (c), we show the accuracy (MSE) of \textsc{FedHLCR} for various fractions of selected agents per iteration and choices of learning rate $\gamma$.
Smaller numbers of selected agents cause larger variations in training data distribution in each iteration, thereby causing
greater instability and worse performance; therefore, choosing a proper $\gamma$ is necessary to ensure convergence.
Figure~\ref{fig:synthlcr} (b) shows the convergence processes with respect to different $\gamma$s.
Here, we observe smooth convergence of the algorithm when $\gamma=0.1$.
When $\gamma=0.2$, we obtain a faster convergence.
If $\gamma$ is increased to 0.5, the accuracy increases faster in the beginning and the convergence is no longer guaranteed, and if $\gamma$ becomes even bigger (0.75), we can see the accuracy is highly unstable.
Figure~\ref{fig:synthlcr} (c) illustrates the relationship between $\gamma$ and accuracy with respect to different fraction of selected agents. It is shown that, for each fraction, there exists an optimal value of $\gamma$ for which a good performance can be achieved in a limited number of iterations.

\paragraph{Real data}


We setup a character-recognition experiment on the FEMNIST dataset~\cite{Caldasetal18, Cohenetal17} as packaged in Federated Tensorflow~\cite{tff}.
This dataset stores images of handwritten characters together with an identifier of the writer (agent).
Let us consider building a character recognition system using only a fraction $F$ of writers among the \textsc{FEMNIST} dataset. We use a Convolutional Neural Network (CNN) taken from~\cite{McMahan2017}\
\footnote{The CNN has two 5x5 convolution layers of 32 and 64 channels, each followed with 2x2 max pooling, a fully connected ReLu layer with a size of 512, and a final softmax output layer with a size of either 10, 36 or 62 for \textsc{FEMNISTdigits}, \textsc{FEMNISTnocaps}, and \textsc{FEMNISTfull} respectively.}
which we train on this subset of writers.
The problem we try to solve is to use the pre-trained CNN for seperating digits from letters in the whole FEMNIST dataset.
Binary classification can in practice be tackled using regression on a \{0, 1\} target.
First, for each subset with different fraction ($F=1\%, 5\%, 10\%, 20\%$), we train a CNN model on the subset.
Then, we consider the following cases:
1) Baseline (CNN-only): The $argmax$ of the vector output by the CNN is used to predict the character and decides on the task (letter vs. digit);
2) LR: A Linear Regression is added after the CNN to perform the binary classification task;
and 3) $\textsc{HLCR}_k$: A HLCR with cluster number $k$ is added after the CNN to perform the binary classification task.
%
Here, we run $\textsc{HLCR}_k$ with writers as agents, using either $4$ or $8$ clusters and a single entity per agent.
Finally, we compare the results of ``$\textsc{HLCR}_4$'' and ``$\textsc{HLCR}_8$'' to those of ``Baseline'' and ``LR''.
The results are reported in Table~\ref{tab:calib:femnist}.
We see that the CNN training time on our CPU-only commodity machine increases linearly with the number of selected agents (around $35$ minutes for 1\% agents for 100 iterations), while training the \textsc{HLCR} model after the CNN takes less than six minutes on the whole dataset.

\begin{table*}
    \center
    \caption{\label{tab:calib:femnist}
        AUC on Classification Tasks on FEMNIST.
        \emph{F}(\%) denotes the fraction of data used in the training of the CNN.
        The Computation Time is the elapsed time on our CPU-only commodity machine.
        CNN training is limited to 100 FL rounds using FedAvg and 5 local epochs.
    }
    \begin{tabular}{c|cccc|cccc}
    & \multicolumn{4}{c|}{AUC (\%)} & \multicolumn{4}{c}{Computation Time (min.)}\\
    $F (\%)$ & Baseline &  LR  &  $\textsc{HLCR}_4$ &  $\textsc{HLCR}_8$ &  Baseline &  LR &  $\textsc{HLCR}_4$ & $\textsc{HLCR}_8$ \\
    \hline
     1 & 93.55 & 93.60 & 96.47 & 96.51 &  30 &  30.5 &  32.1 &  35.1 \\    
     5 & 94.55 & 94.65 & 97.16 & 97.16 & 168 & 168.5 & 170.7 & 173.0 \\
    10 & 94.70 & 94.77 & 97.21 & 97.25 & 334 & 334.5 & 336.7 & 339.9 \\
    20 & 94.74 & 94.77 & 97.17 & 97.20 & 773 & 773.5 & 776.0 & 778.0
    \end{tabular}
\end{table*}

%% file: conclusion.tex
In this paper, we proposed a novel probabilistic model to deal with noisy, hierarchical and tabular datasets, named HLCR.
By applying the Collapsed Gibbs Sampling technique, we efficiently inferred an HLCR model and theoretically guaranteed its convergence.
Furthermore, we provided an HLCR algorithm in Federated Learning, called FedHLCR, for preserving the privacy of agents.
Finally, the experimental results showed that the algorithm offers both fast convergence and good robustness to overfitting even in non-IID datasets.
The immediate future work is to extend the model of each cluster in FedHLCR to a nonlinear model.

%% file: broader_impact.tex
We consider the regression problem for the data held by each agent in Federated Learning. We assume that each agent contains data (events) belonging to different entities, and such agent-entity-event hierarchical data widely exist in various realistic application scenarios, such as smartphones, IoT devices and medical data. In order to analyze and make prediction using such hierarchical data, we propose a Federated Hierarchical Latent Class Regression (FedHLCR) model, which is a mixture of linear regression, so it has richer expressiveness than simple linear regression. With its hierarchical mixture approach, FedHLCR can handle hierarchical data fairly well and given its similar complexity compared to complex models, such as Deep Neural Networks, it can be more efficiently trained by Collapsed Gibbs Sampling. We expect that it will be widely used in various applications. On the other hand, each agent in FedHLCR sends its local training result to the server. This information may cause a privacy risk when an agent only contains very few data. We suggest that FedHLCR application protects the personal information by introducing other privacy-enhancing technologies, such as differential privacy, into FedHLCR.

%% file: appendix.tex
\section{Appendix}

\subsection{Proofs}

\begin{refproof}[Proof of Theorem \ref{thm.likelihood}]
Although $y_{ijn}$s here are NOT independent, generally we have
\begin{equation}
p(\mathbf{y}_{ij} | z_{ij}^{(new)}=k, \mathbf{X}_{ij}, \mathbf{X}_{\setminus ij}^{(k)}, \mathbf{y}_{\setminus ij}^{(k)}, \delta, \sigma) = \prod_{n=1}^{N_{ij}} p(y_{ijn} | \mathbf{y}_{ij[n-1]}, z_{ij}^{(new)}=k, \mathbf{X}_{ij}, \mathbf{X}_{\setminus ij}^{(k)}, \mathbf{y}_{\setminus ij}^{(k)}, \delta, \sigma).
\end{equation}

Let $\mathbf{X}_{ij[n]}$ and $\mathbf{y}_{ij[n]}$ denote $(n \times F)$-matrix $( \mathbf{x}_{ij1}, \mathbf{x}_{ij2}, \cdots, \mathbf{x}_{ijn} )^\top$ and $n$-vector $( y_{ij1}, y_{ij2}, \cdots, y_{ijn} )^\top$. Then, we have $\sum_{m=1}^n \mathbf{x}_{ijm} \mathbf{x}_{ijm}^\top = \mathbf{X}_{ij[n]}^\top \mathbf{X}_{ij[n]}$ and $\sum_{m=1}^n \mathbf{x}_{ijm} y_{ijm} = \mathbf{X}_{ij[n]}^\top \mathbf{y}_{ij[n]}$. With Bayes' theorem,
\begin{eqnarray}
p_{ijn} & := & p(y_{ijn} | \mathbf{y}_{ij[n-1]}, z_{ij}^{(new)}=k, \mathbf{X}_{ij}, \mathbf{X}_{\setminus ij}^{(k)}, \mathbf{y}_{\setminus ij}^{(k)}, \delta, \sigma) \\
 & \propto & p(\mathbf{y}_{ij[n]}, \mathbf{y}_{\setminus ij}^{(k)} | z_{ij}^{(new)}=k, \mathbf{X}_{ij}, \mathbf{X}_{\setminus ij}^{(k)}, \delta, \sigma) \\
 & = & \int p(\mathbf{w}_k, \mathbf{y}_{ij[n]}, \mathbf{y}_{\setminus ij}^{(k)} | z_{ij}^{(new)}=k, \mathbf{X}_{ij[n]}, \mathbf{X}_{\setminus ij}^{(k)}, \delta, \sigma) d\mathbf{w}_k \\
 & = & \int p(\mathbf{w}_k | \delta) p(\mathbf{y}_{\setminus ij}^{(k)} | \mathbf{X}_{\setminus ij}^{(k)}, \mathbf{w}_k, \sigma) p(\mathbf{y}_{ij[n]} | z_{ij}^{(new)}=k, \mathbf{X}_{ij[n]}, \mathbf{w}_k, \sigma) d\mathbf{w}_k \\
 & \propto & \int \exp{ \left\{ -\frac{1}{2\delta^2} \mathbf{w}_k^\top \mathbf{w}_k \right\} } \exp{ \left\{ -\frac{1}{2\sigma^2} (\mathbf{y}_{\setminus ij}^{(k)} - \mathbf{X}_{\setminus ij}^{(k)} \mathbf{w}_k)^\top (\mathbf{y}_{\setminus ij}^{(k)} - \mathbf{X}_{\setminus ij}^{(k)} \mathbf{w}_k) \right\} } \\
 & & \exp{ \left\{ - \frac{1}{2\sigma^2} (\mathbf{y}_{ij[n]} - \mathbf{X}_{ij[n]} \mathbf{w}_k)^\top (\mathbf{y}_{ij[n]} - \mathbf{X}_{ij[n]} \mathbf{w}_k) \right\} } d\mathbf{w}_k \\
 & \propto & \int \exp{\left\{ -\frac{1}{2} \left( \mathbf{w}_k^\top \mathbf{A}_n \mathbf{w}_k - 2 \mathbf{b}_n^\top \mathbf{w}_k + \frac{1}{\sigma^2} y_{ijn}^2 \right) \right\}} d\mathbf{w}_k,
\end{eqnarray}
where
\begin{eqnarray}
\mathbf{A}_n & = & \frac{1}{\delta^2} \mathbf{I} + \frac{1}{\sigma^2} \left( \mathbf{X}_{\setminus ij}^{(k)\top} \mathbf{X}_{\setminus ij}^{(k)} + \sum_{m=1}^n \mathbf{x}_{ijm} \mathbf{x}_{ijm}^\top \right) \\
\mathbf{b}_n & = & \frac{1}{\sigma^2} \left( \mathbf{X}_{\setminus ij}^{(k)\top} \mathbf{y}_{\setminus ij}^{(k)} + \sum_{m=1}^n \mathbf{x}_{ijm} y_{ijm} \right).
\end{eqnarray}
By marginalizing $\mathbf{w}_k$ out, we can get
\begin{eqnarray}
p_{ijn} & \propto & \exp{\left\{ -\frac{1}{2} \left(\frac{1}{\sigma^2} y_{ijn}^2 - \mathbf{b}_n^\top \mathbf{A}_n^{-1} \mathbf{b}_n \right) \right\}} \\
 & \propto & \exp{\left\{ -\frac{1}{2} \left(\frac{1}{\sigma^2} y_{ijn}^2 - (\mathbf{b}_{n-1} + \frac{1}{\sigma^2} \mathbf{x}_{ijn} y_{ijn})^\top \mathbf{A}_n^{-1} (\mathbf{b}_{n-1} + \frac{1}{\sigma^2} \mathbf{x}_{ijn} y_{ijn}) \right) \right\}} \\
 & \propto & \exp{\left\{ -\frac{1 - \frac{1}{\sigma^2} \mathbf{x}_{ijn}^\top \mathbf{A}_n^{-1} \mathbf{x}_{ijn}}{2 \sigma^2} \left( y_{ijn} - \frac{\mathbf{b}_{n-1}^\top \mathbf{A}_n^{-1} \mathbf{x}_{ijn}}{1 - \frac{1}{\sigma^2} \mathbf{x}_{ijn}^\top \mathbf{A}_n^{-1} \mathbf{x}_{ijn}} \right)^2 \right\}}.
\end{eqnarray}
In other words, conditional distribution of $y_{ijn}$ obeys normal distribution, i.e.,
\begin{equation}\label{eq:normal.y}
y_{ijn} | z_{ij}^{(new)}=k, \mathbf{x}_{ijn}, \mathbf{X}_{\setminus ij}^{(k)}, \mathbf{y}_{\setminus ij}^{(k)}, \delta, \sigma \sim \mathcal{N} \left( \frac{\mathbf{b}_{n-1}^\top \mathbf{A}_n^{-1} \mathbf{x}_{ijn}}{1 - \frac{1}{\sigma^2} \mathbf{x}_{ijn}^\top \mathbf{A}_n^{-1} \mathbf{x}_{ijn}}, \frac{\sigma^2}{1 - \frac{1}{\sigma^2} \mathbf{x}_{ijn}^\top \mathbf{A}_n^{-1} \mathbf{x}_{ijn}} \right).
\end{equation}
We have $\mathbf{A}_n = \mathbf{A}_{n-1} + \frac{1}{\sigma^2} \mathbf{x}_{ijn} \mathbf{x}_{ijn}^\top$. With the Woodbury matrix identity,
\begin{equation}
\mathbf{A}_n^{-1} = \mathbf{A}_{n-1}^{-1} - \frac{1}{\sigma^2} \mathbf{A}_{n-1}^{-1} \mathbf{x}_{ijn} (1 + \frac{1}{\sigma^2} \mathbf{x}_{ijn}^\top \mathbf{A}_{n-1}^{-1} \mathbf{x}_{ijn})^{-1} \mathbf{x}_{ijn}^\top \mathbf{A}_{n-1}^{-1}.
\end{equation}
Substituting it into (\ref{eq:normal.y}), we obtain
\begin{eqnarray}
\frac{\mathbf{b}_{n-1}^\top \mathbf{A}_n^{-1} \mathbf{x}_{ijn}}{1 - \frac{1}{\sigma^2} \mathbf{x}_{ijn}^\top \mathbf{A}_n^{-1} \mathbf{x}_{ijn}} & = & \mathbf{b}_{n-1}^\top \mathbf{A}_{n-1}^{-1} \mathbf{x}_{ijn}, \\
\frac{\sigma^2}{1 - \frac{1}{\sigma^2} \mathbf{x}_{ijn}^\top \mathbf{A}_n^{-1} \mathbf{x}_{ijn}} & = & \sigma^2 + \mathbf{x}_{ijn}^\top \mathbf{A}_{n-1}^{-1} \mathbf{x}_{ijn}. \\
\end{eqnarray}
Therefore, we come to this conclusion.
\end{refproof}

%
%

\begin{refproof}[Proof of Theorem \ref{thm.prediction}]
Let us evaluate the distribution
\begin{equation}
p(y_{ij(N_{ij}+1)} | \mathbf{x}_{ij(N_{ij}+1)}, z_{ij}, \mathbf{X}, \mathbf{y}, \delta, \sigma).
\end{equation}

\begin{eqnarray}
 & & p(y_{ij(N_{ij}+1)} | \mathbf{x}_{ij(N_{ij}+1)}, z_{ij}, \mathbf{X}, \mathbf{y}, \delta, \sigma) \\
 & = & p(y_{ij(N_{ij}+1)} | \mathbf{x}_{ij(N_{ij}+1)}, \mathbf{X}^{(z_{ij})}, \mathbf{y}^{(z_{ij})}, \delta, \sigma) \\
 & \propto & p(y_{ij(N_{ij}+1)}, \mathbf{y}^{(z_{ij})} | \mathbf{x}_{ij(N_{ij}+1)}, \mathbf{X}^{(z_{ij})}, \delta, \sigma) \\
 & = & \int p(\mathbf{w}_{z_{ij}}, y_{ij(N_{ij}+1)}, \mathbf{y}^{(z_{ij})} | \mathbf{x}_{ij(N_{ij}+1)}, \mathbf{X}^{(z_{ij})}, \delta, \sigma) d\mathbf{w}_{z_{ij}} \\
 & = & \int p(\mathbf{w}_{z_{ij}} | \delta) p(\mathbf{y}^{(z_{ij})} | \mathbf{X}^{(z_{ij})}, \mathbf{w}_{z_{ij}}, \sigma) p(y_{ij(N_{ij}+1)} | \mathbf{x}_{ij(N_{ij}+1)}, \mathbf{w}_{z_{ij}}, \sigma) d\mathbf{w}_{z_{ij}} \\
 & \propto & \int \exp{ \left\{ -\frac{1}{2\delta^2} \mathbf{w}_{z_{ij}}^\top \mathbf{w}_{z_{ij}} \right\} } \\
 & & \exp{ \left\{ -\frac{1}{2\sigma^2} (\mathbf{y}^{(z_{ij})} - \mathbf{X}^{(z_{ij})} \mathbf{w}_{z_{ij}})^\top (\mathbf{y}^{(z_{ij})} - \mathbf{X}^{(z_{ij})} \mathbf{w}_{z_{ij}}) \right\} } \\
 & & \exp{ \left\{ - \frac{1}{2\sigma^2} (y_{ij(N_{ij}+1)} - \mathbf{x}_{ij(N_{ij}+1)}^\top \mathbf{w}_{z_{ij}})^2 \right\} } d\mathbf{w}_{z_{ij}} \\
 & \propto & \int \exp{\left\{ -\frac{1}{2} \left( \mathbf{w}_{z_{ij}}^\top \mathbf{A} \mathbf{w}_{z_{ij}} - 2 \mathbf{b}^\top \mathbf{w}_{z_{ij}} + \frac{1}{\sigma^2} y_{ij(N_{ij}+1)}^2 \right) \right\}} d\mathbf{w}_{z_{ij}},
\end{eqnarray}
where
\begin{eqnarray}
\mathbf{A} & = & \frac{1}{\delta^2} \mathbf{I} + \frac{1}{\sigma^2} \left( \mathbf{X}^{(z_{ij})\top} \mathbf{X}^{(z_{ij})} + \mathbf{x}_{ij(N_{ij}+1)} \mathbf{x}_{ij(N_{ij}+1)}^\top \right) \\
\mathbf{b} & = & \frac{1}{\sigma^2} \left( \mathbf{X}^{(z_{ij})\top} \mathbf{y}^{(z_{ij})} + \mathbf{x}_{ij(N_{ij}+1)} y_{ij(N_{ij}+1)} \right).
\end{eqnarray}
Therefore, it is straightforward that
\begin{eqnarray}
\mathbf{A} & = & \mathbf{D}^{(z_{ij})} + \frac{1}{\sigma^2} \mathbf{x}_{ij(N_{ij}+1)} \mathbf{x}_{ij(N_{ij}+1)}^\top \\
\mathbf{b} & = & \mathbf{c}^{(z_{ij})} + \frac{1}{\sigma^2} \mathbf{x}_{ij(N_{ij}+1)} y_{ij(N_{ij}+1)}.
\end{eqnarray}
By marginalizing $\mathbf{w}_k$ out, we can get
\begin{eqnarray}
p(y_{ij(N_{ij}+1)} | \mathbf{x}_{ij(N_{ij}+1)}, z_{ij}, \mathbf{X}, \mathbf{y}, \delta, \sigma) & \propto & \exp{\left\{ -\frac{1}{2} \left(\frac{1}{\sigma^2} y_{ij(N_{ij}+1)}^2 - \mathbf{b}^\top \mathbf{A}^{-1} \mathbf{b} \right) \right\}} \\
 & \propto & \exp{\left\{ -\frac{\tau}{2} \left( y_{ij(N_{ij}+1)} - \mu \right)^2 \right\}}.
\end{eqnarray}
where
\begin{eqnarray}
\mu & = & \frac{{\mathbf{c}^{(z_{ij})}}^\top \mathbf{A}^{-1} \mathbf{x}_{ij(N_{ij}+1)}}{1 - \frac{1}{\sigma^2} \mathbf{x}_{ij(N_{ij}+1)}^\top \mathbf{A}^{-1} \mathbf{x}_{ij(N_{ij}+1)}}. \label{eq.prediction.mean}
\end{eqnarray}
It can be shown that $y_{ij(N_{ij}+1)}$ obeys a normal distribution with expected value $\mu$ and precision $\tau$; i.e., $\mathcal{N}(\mu, 1/\tau)$.
With the Woodbury matrix identity,
\begin{eqnarray}
\mathbf{A}^{-1} & = & (\mathbf{D}^{(z_{ij})} + \frac{1}{\sigma^2} \mathbf{x}_{ij{N_{ij}+1}} \mathbf{x}_{ij(N_{ij}+1)}^\top)^{-1} \\
 & = & \mathbf{H}^{(z_{ij})} - \mathbf{H}^{(z_{ij})} \mathbf{x}_{ij(N_{ij}+1)} (\sigma^2 + \mathbf{x}_{ij(N_{ij}+1)}^\top \mathbf{H}^{(z_{ij})} \mathbf{x}_{ij(N_{ij}+1)})^{-1} \mathbf{x}_{ij(N_{ij}+1)}^\top \mathbf{H}^{(z_{ij})}.
\end{eqnarray}
Setting $u = {\mathbf{c}^{(z_{ij})}}^\top \mathbf{H}^{(z_{ij})} \mathbf{x}_{ij(N_{ij}+1)}$ and $v = \mathbf{x}_{ij(N_{ij}+1)}^\top \mathbf{H}^{(z_{ij})} \mathbf{x}_{ij(N_{ij}+1)}$, we have
\begin{eqnarray}
{\mathbf{c}^{(z_{ij})}}^\top \mathbf{A}^{-1} \mathbf{x}_{ij(N_{ij}+1)} & = & u - u (\sigma^2 + v)^{-1} v, \\
\mathbf{x}_{ij(N_{ij}+1)}^\top \mathbf{A}^{-1} \mathbf{x}_{ij(N_{ij}+1)} & = & v - v (\sigma^2 + v)^{-1} v.
\end{eqnarray}
Substituting them into the expected value of normal distribution in (\ref{eq.prediction.mean}),
\begin{equation}
\mu = \frac{u - u (\sigma^2 + v)^{-1} v}{1 - \frac{1}{\sigma^2} (v - v (\sigma^2 + v)^{-1} v)} = u = {\mathbf{c}^{(z_{ij})}}^\top \mathbf{H}^{(z_{ij})} \mathbf{x}_{ij(N_{ij}+1)}.
\end{equation}
\end{refproof}